\def\year{2018}\relax
\documentclass[letterpaper]{article} 
\usepackage{aaai18modif}  
\usepackage{times}  
\usepackage{helvet}  
\usepackage{courier}  
\usepackage{url}  
\usepackage{graphicx}  

\usepackage{amsmath, amsthm, amssymb,array,graphicx}

\theoremstyle{remark}

\begin{document}
%
\title{ChemGAN challenge for drug discovery: can AI reproduce natural chemical diversity?}
\author{Mostapha Benhenda\footnote{Startcrowd, Online AI Lab. Correspondence: mostaphabenhenda@gmail.com}}
\date{\today}

\maketitle
\begin{abstract}
Generating molecules with desired chemical properties is important for drug discovery. The use of generative neural networks is promising for this task. However, from visual inspection, it often appears that generated samples lack diversity. In this paper, we quantify this internal chemical diversity, and we raise the following challenge: can a nontrivial AI model reproduce natural chemical diversity for desired molecules? To illustrate this question, we consider two generative models: a Reinforcement Learning model and the recently introduced ORGAN. Both fail at this challenge. We hope this challenge will stimulate research in this direction.
\end{abstract}

\section{Introduction}

Drug discovery is like finding a needle in a haysack. The chemical space of potential drugs contains more than $10^{60}$ molecules. Moreover, testing a drug in a medical setting is time-consuming and expensive. Getting a drug to market can take up to 10 years and cost \$2.6 billion \cite{dimasi2016innovation}. In this context, computer-based methods are increasingly employed to accelerate drug discovery and reduce development costs. 

In particular, there is a growing interest in AI-based generative models. Their goal is to generate new lead compounds \textit{in silico}, such that their medical and chemical properties are predicted in advance. Examples of this approach include Variational Auto-Encoders \cite{gomez2016automatic}, Adversarial Auto-Encoders \cite{kadurin2017cornucopia,kadurin2017drugan}, Recurrent Neural Networks and Reinforcement Learning \cite{jaques2017sequence,segler2017generating,olivecrona2017molecular}, eventually in combination with Sequential Generative Adversarial Networks \cite{organ,benjamin_carlos_l._alan_2017}.

However, research in this field often remains at the exploratory stage: generated samples are sometimes evaluated only visually, or with respect to metrics that are not the most relevant for the actual drug discovery process. 

Rigorous evaluation would be particularly welcome regarding the \textbf{internal chemical diversity} of the generated samples. Generating a chemically diverse stream of molecules is important, because drug candidates can fail in many unexpected ways, later in the drug discovery pipeline. 

Based on visual inspection, \cite[p. 8]{jaques2017sequence} reports that their Reinforcement Learning (RL) generative model tends to produce simplistic molecules. On the other hand, \cite[p.6, p.8]{organ} argues that their Objective-Reinforced Generative Adversarial Network (ORGAN) generates less repetitive and less simplistic samples than RL. However, their argument is also based on visual inspection and therefore, it remains subjective: our own visual inspection of the ORGAN-generated samples (available on the ORGAN Github: 

\url{https://github.com/gablg1/ORGAN/tree}

\url{/master/results/mol_results}) rather suggests that ORGAN produces molecules as repetitive and as simplistic as RL.

In this paper, we introduce a metric that quantifies the \textit{internal chemical diversity} of the model output. We also submit a challenge: 

\bigskip
\textbf{Challenge:}
Is it possible to build a non-trivial generative model, with (part of) its output satisfying a non-trivial chemical property, such that the internal chemical diversity of this output is at least equal to the diversity found in nature for the same kind of molecules?

\bigskip

To illustrate this challenge, we compare RL and ORGAN generative models, with respect to the following chemical properties:

\begin{enumerate}

\item Being \textbf{active against the dopamine receptor D2}. The dopamine D2 receptor is the main receptor for all antipsychotic drugs (schizophrenia, bipolar disorder...). 
\item \textbf{Druglikeness} as defined in \cite{organ}. We are interested in this property because we can use experimental results in \cite{organ} to facilitate discussion. However, the notion of druglikeness in \cite{organ} is different from the notion of Quantitative Estimation of Druglikeness (QED) \cite{bickerton2012quantifying}, which is an index measuring different physico-chemical properties facilitating oral drug action. 

Here, druglikeness is the arithmetic mean of the solubility (normalized logP), novelty (which equals 1 if the output is outside of the training set, 0.3 if the output is a valid SMILES in the training set, and 0 if the output is not a valid SMILES), synthesizability (normalized synthetic accessibility score \cite{ertl2009estimation}) and conciseness (a measure of the difference of the length between the generated SMILES and its canonical representation). 

We mention that recently, \cite{benjamin_carlos_l._alan_2017} considers an ORGAN with the QED definition of druglikeness. However, we also performed our own experiments with the QED property, and they did not affect our conclusions.

\end{enumerate}

\section{The metric of internal chemical diversity}

Let $a$ and $b$ be two molecules, and $m_a$ and $m_b$ be their Morgan fingerprints \cite{rogers2010extended}. Their number of common fingerprints is $|m_a\cap m_b|$ and their total number of fingerprints is $|m_a\cup m_b|$.

The Tanimoto-similarity $T_s$ between $a$ and $b$ is defined by:

\[ T_s(a,b)= \frac{|m_a\cap m_b|}{| m_a\cup m_b|} \]

Their Tanimoto-distance is:

\[ T_d(a,b)= 1-T_s(a,b)   \]

We use rdkit implementation \cite{landrum2006rdkit} of this distance.

\subsection{Internal diversity}

We define the \textit{internal diversity} $I$ of a set of molecules $A$ of size $|A|$ to be the average of the Tanimoto-distance $T_d$ of molecules of $A$ with respect to each other. Formally, we have:

\begin{equation}
\label{internaldiversity}
I( A)=  \frac{1}{| A |^2} \sum_{ (x,y) \in A \times A }  T_d(x,y)   
\end{equation}

For a sufficiently large set $A$, any sufficiently large subset $A' \subset A$, sampled with uniform probability, has the same internal diversity as $A$. This property follows from the law of large numbers. We can thus define the internal diversity of a generative model, by computing the internal diversity of a sufficiently large generated sample. This allows to formalize our challenge:

\bigskip
\textbf{Challenge (restatement):} Let $N$ be the molecules observed in nature. Is there a non-trivial generative model $G$ and a non-trivial chemical property $P$ such that:  

\begin{equation}
I(G \cap P) \geq I(N \cap P) \, \,\,\,\,?
\end{equation}

\bigskip

Internal chemical diversity is always smaller than $1$ (because the Tanimoto-distance is smaller than $1$), and it is usually much smaller. That's why we prefer this definition to the Tanimoto-variance of a set of molecules $A$, which is:

\begin{equation}
\label{tanimotovariance}
V( A)=  \frac{1}{| A |^2} \sum_{ (x,y) \in A \times A }  T_d(x,y)^2   
\end{equation}

\subsection{External diversity}

A related notion is \textbf{external diversity}. Let $A_1$ and $A_2$ two sets of molecules. The \textit{relative diversity} $E$ of $A_1,A_2$ is defined by:

\begin{equation}
\label{externaldiversity}
E( A_1,A_2)=  \frac{1}{| A_1 | \times | A_2 | } \sum_{ (x,y) \in A_1 \times A_2}  T_d(x,y)   
\end{equation} 

The \textit{external diversity} of a generative model is defined as the relative diversity between the training set and a sufficiently large generated sample.

External diversity essentially corresponds to the notion of diversity defined in \cite[p.5]{organ}.\footnote{The only difference is that in the definition of \cite[p.5]{organ}, only a random subset of molecules of the training set is considered. For faster computations, we also consider a random subset of the training set (of $3000$ samples).} 

A measure of the Tanimoto similarity between generated and natural molecules is also considered in \cite[figures 7 and 12]{segler2017generating} (and their figure 11 considers the Levenshtein distance between them).

The main insight of our paper is to compare internal diversities of generated and natural molecules respectively, instead of considering the relative diversity between generated and natural molecules (and also, we measure this internal diversity with respect to the subset of molecules satisfying the property of interest). 

We think measuring internal diversity is a good way to quantitatively capture the visually observed fact that generated molecules can be repetitive and simplistic \cite{organ,jaques2017sequence}.

\section{Generative Models}

\subsection{Reinforcement Learning}

As in the case of RL considered in \cite{organ}, the generator $G_{\theta}$ is a LSTM Recurrent Neural Network \cite{hochreiter1997long} parameterized by $\theta$. $G_{\theta}$ generates SMILES (Simplified Molecular-Input Line-Entry System) sequences of length $T$ (eventually padded with "\_" characters), denoted by:

\[ Y_{1:T} = (y_1, ..., y_T ) \]

Let $R(Y_{1:T} )$ be the reward function.

\begin{itemize}

\item For the case of dopamine D2 activity, we take:

\[ R(Y_{1:T} ) = P_{\mbox{active}} (Y_{1:T} ) \]

where $P_{\mbox{active}} (Y_{1:T} )$ is the probability for $Y_{1:T}$ to be D2-active. This probability is given by the predictive model made in \cite{olivecrona2017molecular}\footnote{This reward function is slightly different than the function in \cite{olivecrona2017molecular}, which is: $ -1 + 2 \times P_{\mbox{active}}$.}, and available online at 

\url{https://github.com/MarcusOlivecrona/} 

\url{REINVENT/releases}

\item For the case of druglikeness, we take:

\[ R(Y_{1:T} ) = L(Y_{1:T} )  \]

where $L(Y_{1:T} )$ is the druglikeness of $Y_{1:T}$.

\end{itemize}

The generator $G_{\theta}$ is viewed as a Reinforcement Learning agent: its state $s_t$ is the currently produced
sequence of characters $Y_{1:t}$, and its action $a$ is the next character $y_{t+1}$, which is selected in the alphabet $\mathcal{Y}$. The agent policy
is: $G_{\theta}(y_{t+1}|Y_{1:t})$. It corresponds to the probability to choose $y_{t+1}$ given previous characters $Y_{1:t}$.

Let $Q(s, a)$ be the action-value function. It is the expected reward at state $s$ for taking action $a$ and for following the policy $G_{\theta}$, in order to complete the rest of the sequence. We maximize its expected long-term reward:

\[ J(\theta) = \mathbb{E}[ R(Y_{1:T}) |s_0, \theta] = \sum_{y_1 \in \mathcal{Y}} G_{\theta}(y_1| s_0) Q(s_0, y_1)  \]

For any full sequence $Y_{1:T}$, we have:

\[  Q(s = Y_{1:T-1} , a = y_T ) = R(Y_{1:T} )  \]

For $t<T$, in order to calculate the expected reward $Q$ for $Y_{1:t}$, we perform a $N$-time Monte Carlo search with the
 rollout policy $G_{\theta}$, represented as:

\[ \mbox{MC}(G_{\theta}(Y_{1:t}, N)) = \{Y^1_{1:T}, ..., Y^N_{1:T} \}  \]

where $Y^n_{1:t} = Y_{1:t}$ and $Y^n_{t+1:T}$ is randomly sampled via the policy $G_{\theta}$. 

For $t<T$, $Q$ is given by:

\[  Q(s = Y_{1:t-1}, a = y_t) = \frac{1}{N} \sum_{n=1}^N    R(Y^n_{1:T} ) \]

\subsection{Objective-Reinforced Generative Adversarial Network (ORGAN)}

To obtain an ORGAN, \cite{organ} brings a Character-Aware Neural Language Model \cite{kim2016character} $D_\phi$ parameterized by $\phi$. Basically, $D_\phi$ is a Convolutional Neural Network (CNN) whose output is given to a LSTM. $D_\phi$ is fed with both  training data and data generated by $G_{\theta}$. It plays the role of a \textit{discriminator}, to distinguish between the two:
for a SMILES $Y_{1:T}$, the output $ D_\phi (Y_{1:T})$ is the probability that $Y_{1:T}$ belongs to the training data.

For the case of dopamine D2-activity, the reward function becomes:

\[ R(Y_{1:T} ) = \lambda  D_\phi (Y_{1:T}) + (1-\lambda)  P_{\mbox{active}} (Y_{1:T} )   \]

and for the case of druglikeness:

\[ R(Y_{1:T} ) = \lambda  D_\phi (Y_{1:T}) + (1-\lambda)  L(Y_{1:T} )   \]

where $\lambda \in [0,1]$ is a hyper-parameter. For  $\lambda=0$, we get back the RL case, and for $\lambda=1$, we obtain a Sequential Generative Adversarial Network (SeqGAN) \cite{yu2017seqgan}.

The networks $G_\theta$ and $D_\phi$ are trained adversarially \cite{jurgen1992learning,goodfellow2014generative}, such that the loss function for $D_\phi$ to minimize is given by:

\begin{equation}
\label{lossdiscr}
\min_{\phi}  \mathbb{E}_{Y \sim \mbox{train} }[ \log D_\phi(Y)]+ \mathbb{E}_{Y \sim \mbox{gen} }   [ \log (1-D_\phi(Y))] 
\end{equation}

\section{Experiments}

As in \cite{organ}, we pre-train the models 240 epochs with Maximum Likelihood Estimation (MLE), on a random subset of \textbf{15k molecules from the ZINC database} of 35 million commercially-available compounds for virtual screening, used in drug discovery \cite{sterling2015zinc}. Then we further train the models with RL and ORGAN respectively, for 30 and 60 epochs more. 

\subsection{Dopamine D2 activity}

In table \ref{drd2resultsnew}, we show the proportion of valid SMILES output (Prop. Valid SMILES), the average probability of activity on dopamine D2 (Avg. $P_a$), the average internal diversity (Avg. int. div.), the proportion of molecules with probability of activity greater than $0.8$ (Prop. $P_a> 0.8$), and most importantly, the average internal diversity among samples with probability of activity greater than $0.8$. That's the most important column, because it is related with our open problem. 

The averages are computed over the set of valid SMILES, whereas the proportions are computed over all the generated SMILES (both valid and invalid).

We compute those quantities for a D2-active set of 8324 molecules from ExCAPE-DB \cite{sun2017excape} (which is essentially the training set of the SVM classifier in \cite{olivecrona2017molecular}) (DRD2), for the output of the Reinforcement Learning model after 30 epochs (RL 30) and 60 epochs (RL 60), and for the output of ORGAN with $\lambda = 0.04$ after 30 epochs and 60 epochs (ORGAN-0.04 30, ORGAN-0.04 60) and for $\lambda= 0.5$ after 60 epochs (ORGAN-0.5 60). All those outputs have 32k samples.

\begin{center}
[TABLE \ref{drd2resultsnew} HERE]
\end{center}


\begin{table*}[!htb]
\centering
\begin{tabular}{c*{5}{>{$}c<{$}}}
  & \begin{tabular}[c]{@{}l@{}} Prop. Valid \\  SMILES  \end{tabular}   & \text{Avg. $P_a$}        & \text{Avg. int. div.}       & \text{Prop. $P_a> 0.8$}       & \begin{tabular}[c]{@{}l@{}}Avg. int. div. \\  for $P_a> 0.8$ \end{tabular}      \\
DRD2  &     0.996636    &     0.911519      &     0.089478          & 0.876367     &   0.081972     \\
RL 30  &      0.379844     & 	0.160777      & 0.112864  &  0.018906 &  8.65864e-05	 \\
RL 60  & 0.536	 &    0.389979 & 0.014994 & 0.078438	 &  	0.000775 \\
ORGAN-0.04 30  &  0.425375 & 0.097810 & 0.242544 & 0.013531  & 	0.005826 \\
ORGAN-0.04 60  & 0.604406  & 0.342687	 & 0.028563 &  0.100969  & 0.000170 \\
ORGAN-0.5 60  & 0.264687 & 0.006502 &  0.324884	 &  0.000187	 & 0.0	 \\
\end{tabular}
\caption{Experimental results for probability of D2 activity $P_a$ \label{drd2resultsnew}}
\end{table*}

The most interesting case is RL after 30 epochs. In this case, we can see that increasing the probability of D2 activity is contradictory with keeping diversity. After 30 epochs, internal diversity is still pretty good overall, even higher than the DRD2 diversity baseline.

However, when we only keep the molecules of interest, with $P_a> 0.8$, internal diversity dramatically drops to vanishingly small levels.

For ORGAN-$0.04$, results are mostly analogous to RL. We note that at 30 epochs, diversity for $P_a> 0.8$ is 2 orders of magnitude better than RL 30. However, it still remains one order of magnitude lower than the DRD2 baseline, and at 60 epochs, diversity has dropped to levels similar with RL. 

For ORGAN-$0.5$, learning the D2 property still did not start after 60 epochs. The situation is analogous to the SeqGAN case ($\lambda=1$) described in \cite{organ}: high diversity, but no learning of the objective. In particular, that's why the internal diversity for $P_a> 0.8$ is indetectable: there are only 6 samples satisfying the desired property, among 32k.
 
The intermediate cases between $\lambda=0.04$ and $\lambda=0.5$ are analogous to either of them. It is hard to situate the tipping point, between the cases where training is just slow, and where training will never take off.

Here are 10 samples for ORGAN with $\lambda=0.04$ after 30 epochs, selected such that $P_a> 0.8$ (most diverse case):

\bigskip

CCOCCNC[C@H]1CCCN1CCc1ccsc1

CCCOC[C@H]1Cc2ccccc21

CC[C@H]1CCNCOc2ccccc21

CC[C@H]1CCN(CCc2ccccc2)c1

CCCO[C@@H]1CCN(C)Cc2ccccc21

CCC[C@@H]1CCC[NH+]1CC[C@H]1CCCn1

CC[C@@H]1CCN(CCc2ccccc2)c1 

CC[C@H]1CCN(Cc2ccccc2)c1

CCOC1CCN(CCCNCCCc2ccccc2)c1

CCCN1CCO[C@H]1C[C@@H]1CCOc2ccccc21

\subsection{Druglikeness}

In table \ref{druglikeresults}, we show the proportion of valid SMILES output (prop. Valid SMILES), average druglikeness (Avg. $L$), the average internal diversity (Avg. int. div.), the proportion of molecules with druglikeness greater than $0.8$ (Prop. $L> 0.8$), and most importantly, the average internal diversity among samples with druglikeness greater than $0.8$. Again, that's the most important column, because it is related with our challenge. 

Again, the total averages are computed over the set of valid SMILES, whereas the proportions are computed over all the generated SMILES (both valid and invalid).

We compute those quantities for the training set ZINC of 15k molecules (ZINC), which serves as a baseline, for the output of the Reinforcement Learning model after 200 epochs (RL 200) and for the output of ORGAN with $\lambda = 0.8$ after 200 epochs (ORGAN 200). Those outputs have 6400 samples.

\begin{center}
[TABLE \ref{druglikeresults} HERE]
\end{center}

\begin{table*}[!htb]
\centering
\begin{tabular}{c*{5}{>{$}c<{$}}}
  & \begin{tabular}[c]{@{}l@{}} Prop. Valid \\  SMILES  \end{tabular}   & \text{Avg. $L$}        & \text{Avg. int. div.}       & \text{Prop. $L> 0.8$}       & \begin{tabular}[c]{@{}l@{}}Avg. int. div. \\  for $L> 0.8$ \end{tabular}      \\
ZINC  & 1 & 0.661094 & 0.331222 & 0.020133 &  0.025986 \\
RL 200  &   0.975625   & 0.917358  &    0   &  0.974844     &  0     \\
ORGAN-0.8 200  & 0.943906  & 0.906885  & 0.000151 & 0.940625 &  0.000150 \\
\end{tabular}
\caption{Experimental results for Druglikeness $L$ \label{druglikeresults}}
\end{table*}

Results show that ORGAN indeed improves over RL, since it is able to raise internal diversity to detectable levels. However, ORGAN diversity still remains 2 orders of magnitudes lower than ZINC diversity when $L>0.8$. ORGAN diversity also remains 3 orders of magnitude lower than the total diversity of ZINC, which corresponds to the level of internal diversity to which most eyes are used to. We conclude that both RL and ORGAN for $\lambda=0.8$ fail to generate diverse molecules for this property.

Here are 10 SMILES samples from ORGAN for $\lambda=0.8$ and 200 epochs:

\bigskip

Cc1ccccc1CCSc1ccccc1C

COCCc1ccccc1CCCCCCSC

CCCCCCn1cccc1CCCSC

COCc1ccccc1CCCc1ccccc1CC

CCCC(=O)CCCc1ccccc1CCCc1ccccc1C

COCCNC(=O)CCc1ccccc1

CCOC(=O)CCCCc1ccccc1CCN(C)CCCCc1ccccc1C

COCCCC(=O)CC(C)CCCCc1ccccc1C

CCCC(=O)CSCCC(=O)CCCCC(=O)OC

CCC(=O)COCCCCCCC(=O)CCCSC

\section{Conclusion and future work}

We conclude that both RL and ORGAN fail to match natural chemical diversity for desired molecules, although ORGAN is slightly better than RL. For future work, ORGAN training can be improved by considering 2 distinct problems:

\begin{itemize}
\item The perfect discriminator problem in adversarial training
\item The imbalance between different objectives in Reinforcement Learning
\end{itemize}

\subsection{The perfect discriminator problem}

In ORGAN training, the discriminator $D_\phi$ quickly becomes perfect: it perfectly distinguishes between training data and generated data. In general, this situation is not very good for adversarial learning \cite{arjovsky2017towards}. Here, the discriminator still teaches something to the generator. On average, according to the discriminator, the probability for a generated sample to belong to the training set still remains far from $0$, although always smaller than $0.5$. This probability is transmitted to the generator through the reward function.

However, not being able to 'fool' the discriminator, even in the SeqGAN case of $\lambda=1$ (without any other objective), shows  generator weakness: it shows inability to reproduce a plain druglike dataset like ZINC. Training a SeqGAN properly should be a first step towards improving ORGAN.

To achieve this, it might be possible to take a larger generator, to replace the discriminator loss in equation (\ref{lossdiscr}) with another function (like CramerGAN \cite{bellemare2017cramer}), and to use one-sided label smoothing \cite[p.4]{salimans2016improved}.

The discriminator might also overfit training data. Taking a larger training set could help, we took 15k samples here (less than 1MB), and this is small compared with training sets in Natural Language Processing. On the other hand, datasets in drug discovery rarely exceed 10k molecules, and therefore, it could also be interesting to look in the direction of low-data predictive neural networks \cite{altae2017low}.

Once adversarial training is stabilized, it might be interesting to replace all classifiers in the reward function with discriminators adversarially trained on different datasets. Various desired properties might be instilled into generated molecules with multiple discriminators. This might better transmit the chemical diversity present in the various training sets. 

\subsection{Imbalance in multi-objective RL}

The main issue is the imbalance between the various objectives in the reward function, a problem occurring also in RL. Multi-objective reinforcement learning is a broad topic (for a survey, see \cite{roijers2013survey}). 

A problem here is that with a weighted sum, the agent always focuses on the easiest objective, and ignores harder ones. Moreover, the relative difficulty between objectives evolves over time. For example, the average probability of D2 activity initially grows exponentially, and so this growth is small when this probability is near $0$.

Using time-varying adaptive weights might help. Moreover, those weights might not necessarily be linear: For example, the reward function can be of the form $(x^\lambda+y^\lambda)^{1/\lambda}$, which converges towards $\min(x,y)$ as $\lambda \rightarrow -\infty$. Using an objective function of the form $\min(x,y)$ focuses the generator on the hard objective (but in our experiments, due to the perfect discriminator problem, it did not work).

Morever, in the reward function, a penalty can be introduced for newly generated molecules that are too similar with the generated molecules already having the desired properties.

In any case, the (varying) relative weights between different objectives must be determined automatically, and not through guesswork.
In a drug discovery setting, a molecule must simultaneously satisfy a large number of objectives. For example, for an antipsychotic drug, it is not enough to be active against D2. The molecule must also pass toxicity and druglikeness tests. Moreover, to avoid side-effects, the molecule must not be active with D3, D4, serotonin, or histamine. That's a lot of objectives to include in the reward function.

Finally, there is also further work to improve the definition of internal diversity, in order to exclude trivial solutions (for example, a generative model reproducing the training set can also have high internal diversity). This will facilitate the attribution of financial prizes.

\section{Acknowledgement}

Computations were performed with 2 GPUs Nvidia Tesla M60, available from Microsoft Azure Free Trial.

\end{document}